\pdfoutput=1

\documentclass[11pt]{article}

\usepackage{acl}

\usepackage{times}
\usepackage{latexsym}

\usepackage[T1]{fontenc}

\usepackage[utf8]{inputenc}

\usepackage{microtype}

\usepackage{inconsolata}

\usepackage{inconsolata}
\usepackage{graphicx}
\usepackage{multirow}
\usepackage{subcaption}
\usepackage{caption}
\usepackage{xcolor}
\usepackage{ulem}

\usepackage{booktabs}
\usepackage{xspace}

\newcommand{\kalign}{cross-lingual knowledge alignment\xspace}
\newcommand{\CLKA}{CLiKA\xspace}

\title{Multilingual Pretraining and Instruction Tuning Improve Cross-Lingual Knowledge Alignment, But Only Shallowly}

\newcommand*{\affaddr}[1]{#1} 
\newcommand*{\affmark}[1][*]{\textsuperscript{#1}}
\newcommand*{\email}[1]{\texttt{#1}}

\author{%
Changjiang Gao\affmark[1]\quad Hongda Hu\affmark[1]\quad Peng Hu\affmark[1]\quad Jiajun Chen\affmark[1]\quad Jixing Li\affmark[2]\quad Shujian Huang\affmark[1]\thanks{\ \ Corresponding author}\\
\affaddr{\affmark[1]National Key Laboratory for Novel Software Technology, Nanjing University}\\
\affaddr{\affmark[2]Department of Linguistics
and Translation, City University of Hong Kong}\\
\email{\{gaocj, huhd, hup\}@smail.nju.edu.cn}\qquad
\email{chenjj@nju.edu.cn}\\
\email{jixingli@cityu.edu.hk} \qquad
\email{huangsj@nju.edu.cn}%
}

\begin{document}
\maketitle
\begin{abstract}
Despite their strong ability to retrieve knowledge in English, current large language models show imbalance abilities in different languages. Two approaches are proposed to address this, i.e., multilingual pretraining and multilingual instruction tuning. However, whether and how do such methods contribute to the \kalign inside the models is unknown. In this paper, we propose \CLKA, a systematic framework to assess the \kalign of LLMs in the Performance, Consistency and Conductivity levels, and explored the effect of multilingual pretraining and instruction tuning on the degree of alignment. Results show that: while both multilingual pretraining and instruction tuning are beneficial for \kalign, the training strategy needs to be carefully designed. Namely, continued pretraining improves the alignment of the target language at the cost of other languages, while mixed pretraining affect other languages less. Also, the overall \kalign, especially in the conductivity level, is unsatisfactory for all tested LLMs, and neither multilingual pretraining nor instruction tuning can substantially improve the cross-lingual knowledge conductivity. \footnote{Our code and data are available on \href{https://github.com/RiverGao/CLiKA}{GitHub}.}
\end{abstract}

\section{Introduction}

The language imbalance of modern NLP systems has long been discussed \cite{hupkes_taxonomy_2023}. 
Many studies have shown that the performance of current LLMs on English tasks is much higher than non-English tasks \cite{wang_seaeval_2023,zhang_dont_2023}. One possible explanation is
that the knowledge required for completing tasks are learnt mainly from English text. So it could be better retrieved with English than with other languages. 

Recent studies suggest that cross-lingual consistency may be a possible way to narrow the gap between languages \cite{qi2023crosslingual}. 
Ideally, if the knowledge of a fact could be aligned to a ``true'' representation regardless of the language it is described with,  it may be retrieved in any required language, helping the model to generalize across languages. 
In this paper, we refer to this internal mechanism as \textit{\kalign}.

To improve LLMs' performance in non-English languages, two approaches are proposed. The first is \textit{multilingual pretraining}, which add non-English data in the pretraining corpus. 
The second is \textit{multilingual instruction tuning}, i.e., using tasks in different languages or translation-related tasks, to finetune a foundation model \cite{zhang_bayling_2023, zhu2023extrapolating}.
Although these methods do improve LLMs' non-English performance, whether they can bring real \kalign is not well investigated. 


Therefore, the aim of this study is to evaluate the effect of multilingual pretraining and instruction tuning on the \kalign mechanism. However, the evaluation is challenging, because
the improvement of performance may come from the improvement of language ability in a specific language or  the improvement of knowledge alignment.
It is hard to discriminate the effects of the two by performance as the single clue. Furthermore, even if LLMs show higher consistency between two languages, there is still possibility that the knowledge in the two languages are learned correctly but separately. 



To meet this challenge, we propose to assess \kalign systematically, by using 3 deepening levels of measurement:
\begin{itemize}
\setlength{\itemsep}{0pt}
\setlength{\parsep}{0pt}
\setlength{\topsep}{0pt}
\setlength{\partopsep}{0pt}
    \item \textbf{Performance} (PF): achieving similar performance for tasks in different languages; 
    \item \textbf{Consistency} (CT): generating the same output for the same input in different languages; 
    \item \textbf{Conductivity} (CD): retrieving knowledge learned in one language with another.
\end{itemize}


Most previous evaluations of the multilingualism only focus on the PF level \cite{kassner-etal-2021-multilingual, yin-etal-2022-geomlama} and the CT level~\cite{qi2023crosslingual}, but the CD level is closer to the nature of knowledge alignment. 

The evaluation for the CD level is non-trivial, because the successful retrieval of a factual knowledge learned in another language depends not only on the alignment of the knowledge, but also on the basic language ability in the current language. For example,  even if the alignment is correct, retrieving this knowledge in non-English languages such as Japanese is harder than doing it in English,
because the model's basic ability is not as good.

In this regard, we propose a systematic framework, \CLKA (standing for Cross-Lingual Knowledge Alignment), to reveal the effects of different multilingualism. \CLKA considers all three levels of the alignment, with specific metrics for each level. \CLKA includes three comparative settings: \textit{Factual}, \textit{Basic}, and \textit{Fictional}, to further discriminate the effects of language abilities and knowledge alignment in knowledge retrieval.


We apply \CLKA to popular LLMs, including BLOOM \cite{workshop2023bloom}, LLaMA \cite{touvron2023llama, touvron2023llama2}, ChatGPT and their multilingual variants \cite{ cui2023efficient, zhang_bayling_2023}. Our results indicate that:
\begin{itemize}
\setlength{\itemsep}{0pt}
\setlength{\parsep}{0pt}
\setlength{\topsep}{0pt}
    \item The general \kalign of current multilingual LLMs is unsatisfactory. They show imbalanced basic abilities and knowledge PF in English and non-English, and their high knowledge CT comes with low CD, suggesting low cross-lingual knowledge conduction.
    \item Mixed multilingual pretraining improves the basic ability, knowledge PF and CT in multiple languages, while continued pretraining can only improve the knowledge PF in the target language at the cost of other languages. However, both of them cannot improve the knowledge CD of LLMs.
    \item Multilingual instruction tuning improves the basic ability in the target language, and can lower the knowledge PF drop brought by instruction tuning. However, it can hardly improve the knowledge CT and CD.
\end{itemize}

\begin{table*}[ht]
    \centering
    \footnotesize
    \begin{tabular}{ccl}
\hline
Knowledge &
  Dataset &
  \multicolumn{1}{c}{Example} \\ \hline
\multirow{2}{*}{Basic} &
  xCSQA &
  \begin{tabular}[c]{@{}l@{}}\textbf{Question:} The dental office handled a lot of patients who experienced traumatic mouth injury, \\where were these patients coming from? \\ A. town  B. michigan  C. hospital  D. schools  E. office building\\ \textbf{Answer:} C. hospital\end{tabular} \\ \cline{2-3} 
 &
  xCOPA &
  \begin{tabular}[c]{@{}l@{}}\textbf{Premise:} The item was packaged in bubble wrap.\\ \textbf{Question:} What was the cause of this? \\ A. It was fragile.  B. It was small.\\ \textbf{Answer:} A. It was fragile.\end{tabular} \\ \hline
\multirow{2}{*}{Factual} &
  xGeo &
  \begin{tabular}[c]{@{}l@{}}\textbf{Question:} What administrative division of Egypt is Alexandria in?\\ A. Red Sea Governorate  B. Alexandria Governorate  C. Cairo Governorate  D. Emirate of Dubai\\ \textbf{Answer:} B. Alexandria Governorate\end{tabular} \\ \cline{2-3} 
 &
  xPeo &
  \begin{tabular}[c]{@{}l@{}}\textbf{Question:} In what year was Houari Boumediene born? \\ A. 1820  B. 1828  C. 1838  D. 1932\\ \textbf{Answer:} D. 1932\end{tabular} \\ \hline
\multirow{2}{*}{Fictional} &
  Translation &
  \begin{tabular}[c]{@{}l@{}}\textbf{Question:} Could you convert the upcoming English text to German? Tempest Hollow\\ \textbf{Answer:} Sturmhain\end{tabular} \\ \cline{2-3} 
 &
  QA &
  \begin{tabular}[c]{@{}l@{}}\textbf{Question:} Which continent is Tempest Hollow located in?\\ \textbf{Answer:} Vividora\end{tabular} \\ \hline
\end{tabular}
\caption{Example of questions used in each of the testing datasets.}
\label{tab:question-examples}
\end{table*}

\section{Related Work}
\paragraph{Multilingualism of language models.} Due to the ``incident bilingualism'' \cite{briakou-etal-2023-searching} and cross-lingual data sharing \cite{choenni2023languages} in the training corpus, pretrained models, including those English-centered ones, will have multilingual ability and cross-lingual alignment of representations to some extent. On that basis, multilingualism can be further strengthened by adding monolingual data in different languages in the pretraining corpus, resulting in multilingual PLMs such as mBERT \cite{devlin-etal-2019-bert}, mBART \cite{liu-etal-2020-multilingual-denoising}, mT5 \cite{xue-etal-2021-mt5}. However, because the multilingual data is not parallel in these models, their language balance and \kalign is much limited \cite{pires-etal-2019-multilingual}. To address this issue, some work uses supervised parallel data in the pretraining stage to enhance the model's multilingualism, e.g. XLM \cite{conneau2019cross} and BLOOM \cite{workshop2023bloom}. Such method is also used nowadays to train LLMs with better multilingualism, bringing models such as PaLM 2 \cite{anil2023palm} on multiple languages; and ChatGLM \cite{du-etal-2022-glm,zeng2022glm} and Baichuan 2 \cite{yang2023baichuan} mainly focusing on English and Chinese. Also, another popular practise is to fine-tune an English-centered foundation model with translation and instruction data in English and other languages \cite{cahyawijaya2023instructalign}, resulting in models with better translation ability and instruction-following ability in those languages, such as BigTranslate\cite{yang_bigtranslate_2023}, BayLing \cite{zhang_bayling_2023}, x-LLaMA/m-LLaMA \cite{zhu2023extrapolating}, and mFTI \cite{li2023eliciting}. However, despite the performance gain on multilingual benchmarks, the effect of these training methods on \kalign is still to be examined.

\paragraph{Multilingual benchmarks and evaluations.}
Evaluation work is rapidly updating in the field of cross-lingual knowledge alignment. In the PLM era, many cross-lingual NLP benchmark datasets were proposed to test the models' performance on certain aspects in different languages, such as XCOPA \cite{ponti_xcopa_2020} and X-CSQA \cite{lin_common_2021} for commonsense reasoning, and X-FACTR \cite{jiang-etal-2020-x} and multilingual versions of LAMA \cite{kassner-etal-2021-multilingual, yin-etal-2022-geomlama, qi2023crosslingual} for factual knowledge. Some work also tested LLMs' multilingual performance on different NLP tasks \cite{lai2023chatgpt, zhang2023m3exam, ahuja2023mega}.


\paragraph{Knowledge misalignment of language models.}
Previous work have pointed out the imbalance of multilingual pretrained language models (PLMs) \cite{pires-etal-2019-multilingual, qi2023crosslingual}. However, since the ``incident multilingualism'' in pertraining increased a lot for LLMs, this conclusion needs to be reevaluated. \citet{zhang_dont_2023} found that ChatGPT does not perform consistently on tasks in different languages, while exhibiting a translation-like thinking mode. \citet{wang_seaeval_2023} concluded that multilingually-trained models have not attained “balanced multilingual” capabilities, especially on commonsense or factual knowledge. However, they did not differentiate between the two sources of language misalignment. \citet{qi2023crosslingual} further evaluated the cross-lingual consistency of PLMs and the factors affecting it using a rank-based metric. However, an evaluation with all three levels of \kalign considered is yet to be done.

\section{Methods}
Because the cross-lingual knowledge retrieval is affected by both the basic language ability and the three levels of cross-lingual knowledge alignment, our \CLKA framework adopts special testing datasets and metrics to evaluate them separately.

\subsection{Constructing testing data}
We constructed three testing datasets: \textit{Basic}, \textit{Factual}, and \textit{Fictional}. The datasets are all in the multiple choice format for easier evaluation. Also, the data is parallel in the same 10 chosen languages: \texttt{en}, \texttt{de}, \texttt{fr}, \texttt{it}, \texttt{ru}, \texttt{pl}, \texttt{ar}, \texttt{he}, \texttt{zh}, \texttt{ja} (Details are listed in Table \ref{tab:language_countries} in Appendix \ref{sec:appendix-lang}). These languages are chosen because they are widely used, and they enable comparison between and within language families.\footnote{ The datasets will be publicly available along with the publication of this paper.} Table \ref{tab:question-examples} shows the example questions.

\paragraph{\textit{Basic} knowledge.} We consider commonsense as \textit{Basic} knowledge to measure the basic language ability of models in the selected languages. There are two reasons for this. Firstly, commonsense is indispensable for LLMs to generate meaningful answers, so if a model lacks commonsense in some languages, it will be very likely to show poor overall ability in these languages. Secondly, because commonsense is so elementary, that they are unlikely to be explicitly stated in any text in any language \cite{lenat1995cyc}, it is difficult to be learned through short-cuts such as remembering training samples. The two parts of this dataset are:
\begin{itemize}
    \item xCOPA (500 samples per language). COPA \cite{roemmele2011choice} is an English dataset focusing on commonsense causality, where each question is a 1-out-of-2 choice. 
    Although there is already a cross-lingual version of COPA, i.e. XCOPA \cite{ponti_xcopa_2020}, it does not cover the languages considered in this study. Instead, we use DeepL and Google Translate to translate the COPA \texttt{test} set into the other 9 languages. 
    \item xCSQA (1000 samples per language). CSQA \cite{talmor_commonsenseqa_2019} is a challenging English dataset focusing on the semantic relation of common concepts, where each question is a 1-out-of-5 choice. There is also a cross-lingual version of this dataset, i.e. X-CSQA \cite{lin_common_2021}. However, we still use the updated translation of the \texttt{val} set for higher quality.
\end{itemize}

\paragraph{\textit{Factual} knowledge.} This represents the real-life knowledge retrieval scenario, and is deliberately balanced among the tested languages, i.e., the knowledge originates evenly from the 10 languages, and is presented parallelly in all of them. Currently, there is no off-the-shelf dataset that meets the requirements. The dataset contains two parts originating from \href{https://www.wikidata.org/wiki/Wikidata:Main_Page}{Wikidata}:
\begin{itemize}
    \item xGeo (200 samples per language), about cities and the administrative division they belong. For each of the 10 languages, we choose 20 cities in the major countries speaking this language (see Table \ref{tab:language_countries} in Appendix \ref{sec:appendix-lang}), and collect their names, and their administrative divisions' names in the 10 languages with WikiData. Then, we construct a 1-out-of-4 choice for each city-division with 3 randomly picked wrong options. After that, we use templates in the 10 languages (see Appendix \ref{sec:templates}) to write the questions. There are thus 200 samples presented in each language in total, 20 for each original language.
    \item xPeo (180 samples per language), about famous people and their years of birth/death (YOBs/YODs). For each language, we choose 10 famous historical figures from the major countries speaking this language. \footnote{Cases of multiple nationalities and unclear YOBs/YODs are ruled out.} Then, we collect their YOBs, DOBs, and names in the 10 languages from WikiData. We again construct a 1-out-of-4 choice for each person-year with 3 randomly picked wrong options, and then use templates in the 10 languages (see Appendix \ref{sec:templates} for all templates) to write the questions. Specially, the xPeo dataset does not contain historical figures originating from Hebrew, because they are either with multiple nationalities, or are contemporary.
\end{itemize}
This \textit{Factual} dataset can be used to evaluate the PF and CT level of \kalign, but it cannot accurately measure the CD level, because even though we have identified the language origins of the factual knowledge, which language is the knowledge first learned in a model, i.e. the source of conductivity, is unknown. To help measure CD, we then construct the \textit{Fictional} knowledge dataset to test the knowledge conductivity from English to other languages.

\paragraph{\textit{Fictional} knowledge.} This dataset consists of artificial entity-relation knowledge, which do not exist in the training of LLMs, making it possible to observe the learning and transferring of knowledge in different languages. While the entity names and their translations in all the 10 languages are provided for training, the tested relations between entities are only presented in English. Therefore, to answer the non-English relations, the models need to conduct knowledge from English to non-English. The dataset is built by the following steps:
\begin{enumerate}
    \item Names of 400 fictional places and 20 fictional continents are generated in English and translated into the other 9 languages by ChatGPT as the entities. Then, 10 translation templates (see Appendix \ref{sec:templates}) are used to construct the translation training data from English to the other 9 languages (4200 samples per language).
    \item Each place is randomly assigned to a continent to build relations between the entities ($20\times 20=400$ relations). Then, all the relations are filled in 10 English QA templates (see Appendix \ref{sec:templates}) and used as the first part of the training data (4000 samples, English only); Meanwhile, half of the relations are filled in the QA templates in the other 9 languages and respectively used as the second part of the training data (2000 samples per language). Note that in each conductivity experiment, only one non-English presents in the training data.
    \item The other half of the relations excluded in the non-English training data are filled in the first template ("Which continent is \{PLACE\} located in?") and used as the testing data (200 samples per language).
\end{enumerate}
The tested models will be tuned with LoRA \cite{hu2021lora} on the two instruction sets, and the performance on the test set is collected.

\subsection{\CLKA measurements}
The basic ability is measured with the \textit{Basic} knowledge, while the PF and CT alignments are measured with the \textit{Factual} knowledge, and the CD aligment is measured on the \textit{Fictional} knowledge. We design three measurements to score these aspects.

\paragraph{PF: Re-scaled accuracy (RA).} 
The raw accuracy is affected by the question difficulty, and there exists a random baseline for multiple choice questions, making it hard to compare the performance across languages and aspects of model ability. Thus, to focus on the difference across languages and models, we re-scale the accuracy. Suppose the raw accuracy of a model in one language is $A$, we use the accuracy of ChatGPT in English (noted $A_g$) as a reference for difficulty, and the expected accuracy of random choice $A_r=1/n_{\mathrm{choice}}$ as the baseline. The re-scaled accuracy (RA) is calculated as:
\[\mathrm{RA} =\frac{\max\{A-A_r,\ 0\}}{\max\{A_g-A_r, 0\}}\]
Note that RA can exceed 1 as long as the raw accuracy is higher than that of ChatGPT in English. Also, since ChatGPT performs better than random on almost all tested tasks, the denominator is larger than 0. More balanced RAs in English and non-English means better PF alignment.


\paragraph{CT: Correct prediction overlap with English (en-CO).} Similar to \citet{jiang-etal-2020-x}, we use the ratio of consistent and correct predictions between English and another language to measure their CT. Suppose the model gives $n_{X}$ correct answers in language $X$, and among them, $n_{\mathrm{en}X}$ are consistent with its answers in English, the en-CO is:
\[\mathrm{CO}(\mathrm{en}, X)=\frac{n_{\mathrm{en}X}}{n_X}\]
The en-CO ranges from 0 to 1, higher value meaning better CT with English.

\paragraph{CD: Cross-retrieval ratio (XRR).} Suppose $n_{\mathrm{en}}$ is the number of correct answers in English and $n_{\mathrm{en}X}$ is correct in both English and language X on \textit{Fictional} knowledge, 
and $A_r$ is the random accuracy baseline (0.05 for the \textit{Fictional} dataset). The cross-retrieval ratio (XRR) is then calculated as:
\[\mathrm{XRR}(X)=\max\left\{\frac{n_{\mathrm{en}X}}{n_{\mathrm{en}}}-A_r,0\right\}\]
XRR is non-negative and can exceed 1, higher value meaning better CD from English to another language.


\begin{table}[ht]
    \centering
    \footnotesize
    \begin{tabular}{ccc}
    \hline
    Mixed & Cont' & Model             \\ \hline
    N              & N              & LLaMA                  \\
    N              & Y              & Chinese-LLaMA          \\
    Y              & N              & Baichuan2-base, LLaMA2 \\
    Y              & Y              & Chinese-LLaMA2         \\ \hline
    \end{tabular}
    \caption{List of foundation models used in the Chinese case study. "Mixed" stands for mixed pretraining in Chinese, and "Cont'" stands for continued pretraining in Chinese.}
    \label{tab:zh-foundation-models}
\end{table}

\begin{table}[ht]
    \centering
    \footnotesize
    \begin{tabular}{ccc}
    \hline
    \textbf{PT} & \textbf{FT} & \textbf{Model}              \\ \hline
    N           & N           & Alpaca                      \\
    N           & Y           & BayLing                     \\
    Y           & N           & LLaMA2-Chat                 \\
    Y           & Y           & Vicuna v1.5                 \\ \hline
    \end{tabular}
    \caption{List of instruction-tuned LLMs used in the Chinese case study. "PT" stands for whether the model has Chinese pretraining, and "FT" stands for whether the model has Chinese instruction tuning.}
    \label{tab:zh-llms}
\end{table}

\begin{figure*}[ht]
\centering
    \begin{subfigure}[b]{5cm}
    \centering
    \includegraphics[width=5cm]{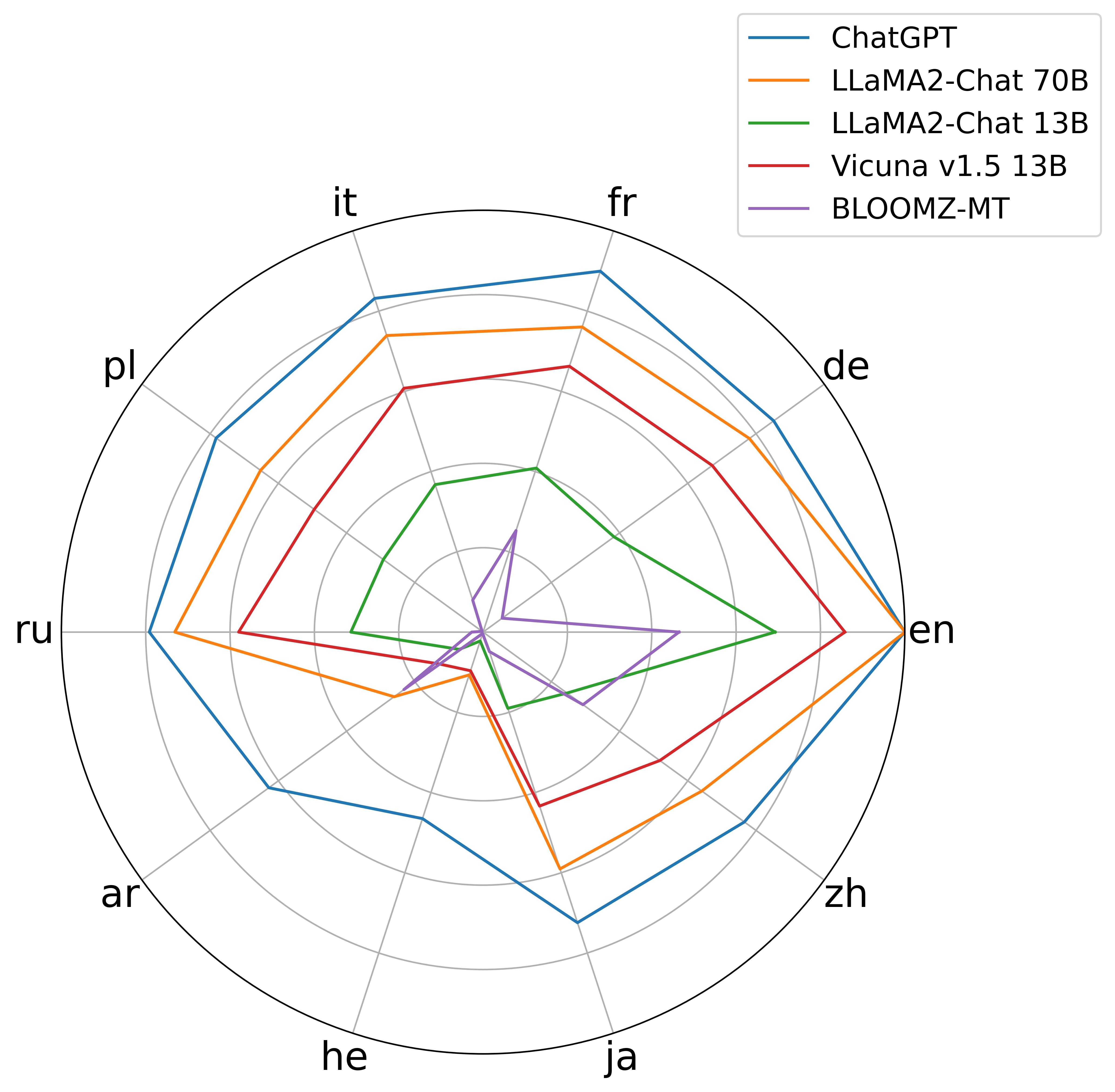}
    \caption{\label{fig:basic-ra}}
    \end{subfigure}
\quad
    \begin{subfigure}[b]{5cm}
    \centering
    \includegraphics[width=5cm]{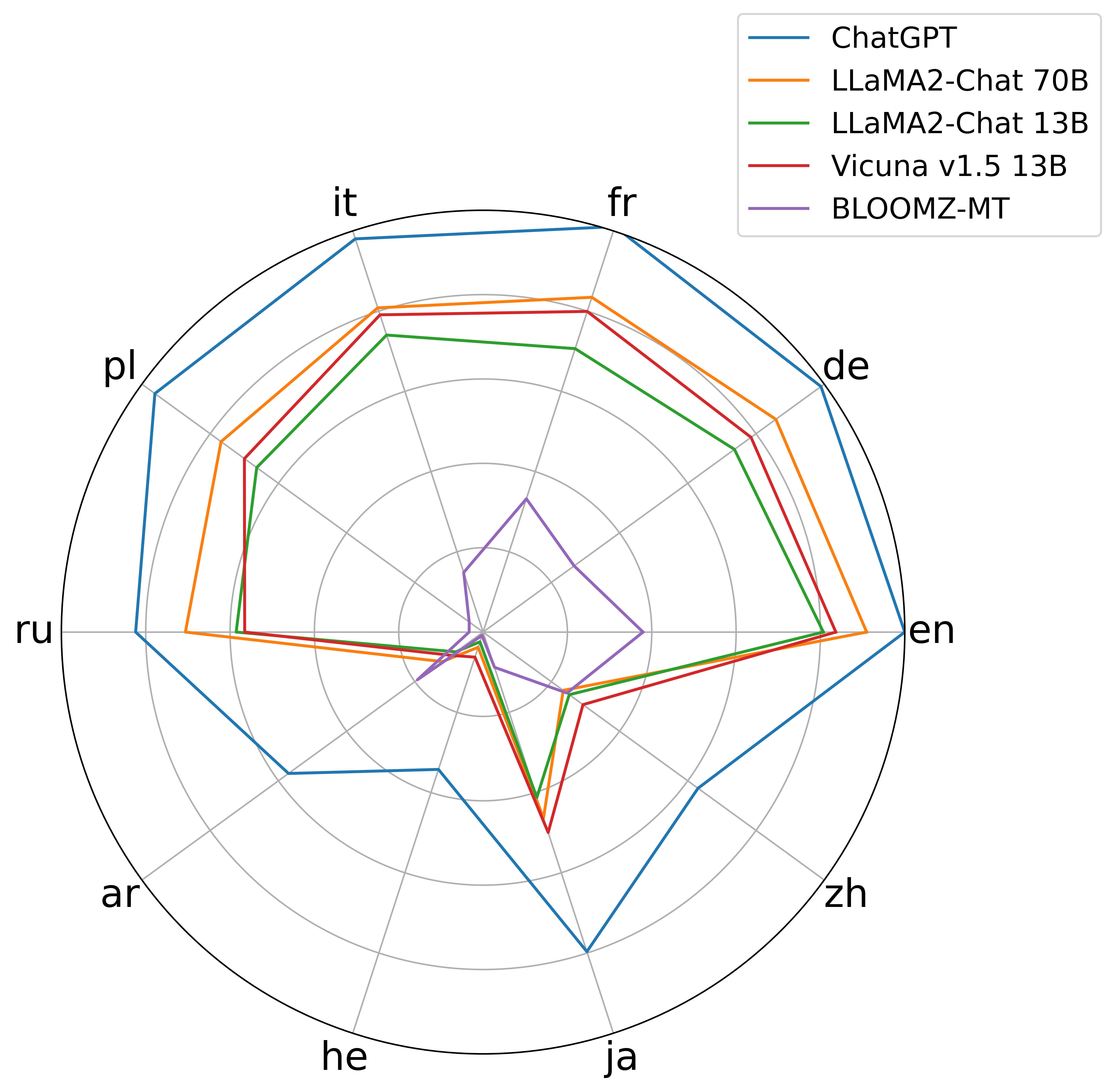}
    \caption{\label{fig:realistic-ra}}
    \end{subfigure}
\quad
    \begin{subfigure}[b]{5cm}
    \centering
    \includegraphics[width=5cm]{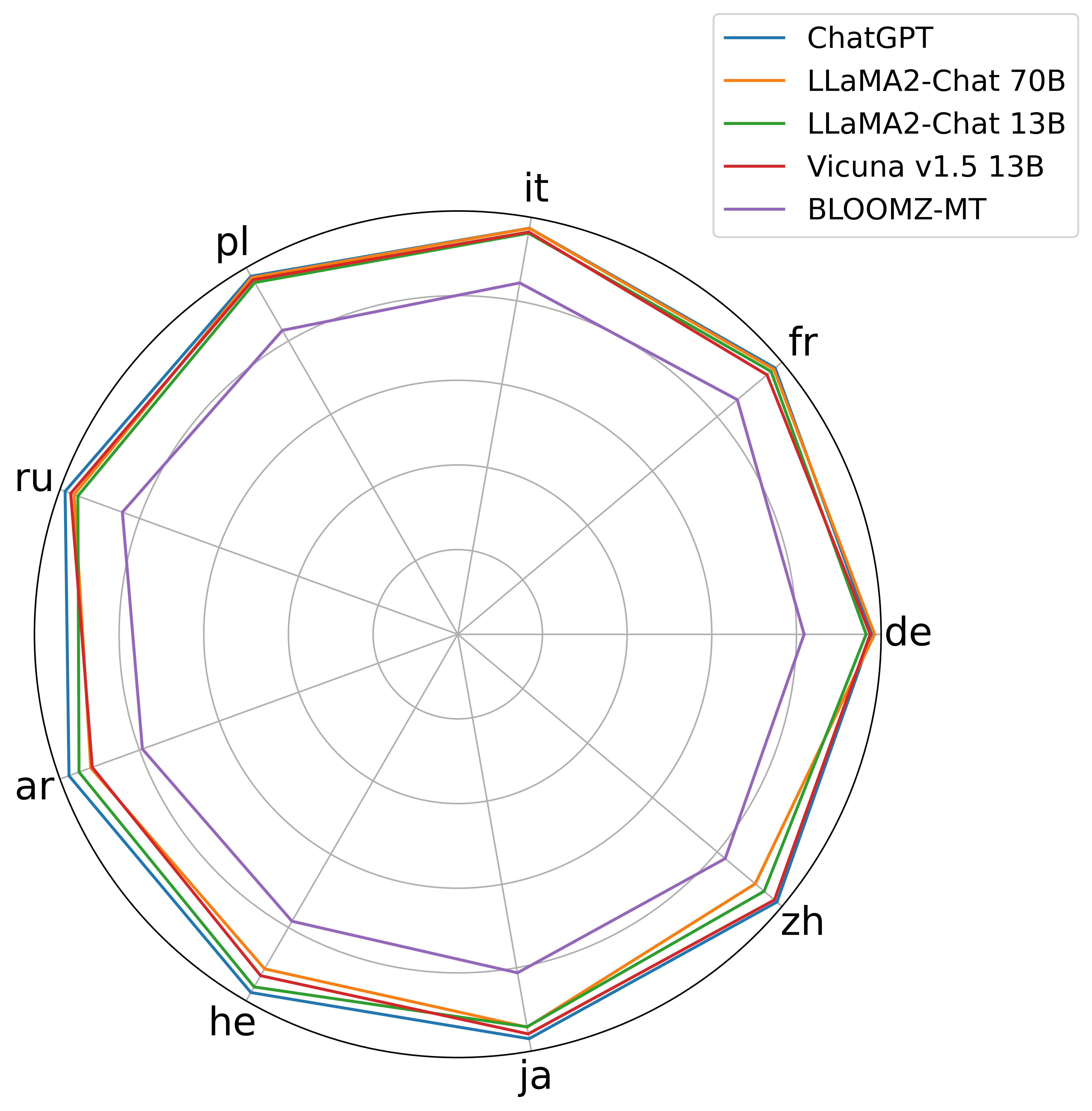}
    \caption{\label{fig:realistic-enco}}
    \end{subfigure}
\caption{Results of the general cross-lingual knowledge alignment evaluation. The outer circle of the radar graphs is 1.0 and the center is 0.0, and each circle represents a 0.2 span. \subref{fig:basic-ra} The RA scores on the \textit{Basic} knowledge (the mean of xCSQA and xCOPA scores; \subref{fig:realistic-ra} The RA scores on the \textit{Factual} knowledge (the mean of xGeo and xPeo scores); \subref{fig:realistic-enco} The en-CO scores on the \textit{Factual} knowledge (the mean of xGeo and xPeo scores).}
\label{fig:general-results}
\end{figure*}

\begin{table*}[ht]
    \centering
    \footnotesize
    \begin{tabular}{cccccccccc}
\hline
Model &
  de &
  fr &
  it &
  pl &
  ru &
  ar &
  he &
  ja &
  zh \\ \hline
LLaMA2-Chat 13B &
  {\color[HTML]{373C43} .0800} &
  {\color[HTML]{373C43} .1200} &
  {\color[HTML]{373C43} .0900} &
  {\color[HTML]{373C43} .1050} &
  {\color[HTML]{F54A45} .0050} &
  {\color[HTML]{F54A45} .0003} &
  {\color[HTML]{F54A45} .0000} &
  {\color[HTML]{F54A45} .0000} &
  {\color[HTML]{F54A45} .0000} \\
Vicuna v1.5 13B &
  {\color[HTML]{373C43} .1309} &
  {\color[HTML]{373C43} .0200} &
  {\color[HTML]{F54A45} .0050} &
  {\color[HTML]{F54A45} .0000} &
  {\color[HTML]{373C43} .0800} &
  {\color[HTML]{373C43} .0800} &
  {\color[HTML]{F54A45} .0000} &
  {\color[HTML]{373C43} .0800} &
  {\color[HTML]{F54A45} .0000} \\
BLOOMZ-7B1-MT &
  {\color[HTML]{373C43} .0200} &
  {\color[HTML]{373C43} .1350} &
  {\color[HTML]{373C43} .0350} &
  {\color[HTML]{373C43} .0656} &
  {\color[HTML]{F54A45} .0050} &
  {\color[HTML]{F54A45} .0000} &
  {\color[HTML]{F54A45} .0000} &
  {\color[HTML]{373C43} .0100} &
  {\color[HTML]{F54A45} .0000} \\ \hline
\end{tabular}
\caption{The XRR scores of representative LLMs on the \textit{Fictional} knowledge. Scores below 0.01 (less than 1\% above-random-accuracy) are colored red.}
\label{tab:general-xrr}
\end{table*}

\section{Experiment Settings}

\subsection{Models}
Our \CLKA analysis of \kalign is two-fold. First, we assess the basic language ability and \kalign of popular multilingual LLMs among all tested languages; then, we examine the effect of multilingual pretraining and instruction tuning on their basic language ability and \kalign,  taking Chinese as a representative for high-resource, non-English language.

The \textbf{popular multilingual LLMs} we selected are ChatGPT (called with the OpenAI API \texttt{gpt-3.5-turbo} in Octorber, 2023), LLaMA 2-Chat 70B and 13B \cite{touvron2023llama2}, Vicuna v1.5 13B \cite{vicuna2023} and BLOOMZ-7B1-MT \cite{workshop2023bloom}.

The models for \textbf{ the Chinese case study} include foundation models (Table \ref{tab:zh-foundation-models}) and LLMs (Table \ref{tab:zh-llms}) that allows comparison of having or not having continued/mixed pretraining and instruction tuning in Chinese (See their information in Appendix \ref{sec:appendix-intro-models}).

\subsection{Assisted inference}
Because some of the models tested are not instruction tuned, directly given the question and options, they may not provide a valid choice. To help them inference, we: (1) use an in-context demonstration for every question to inform them the correct answering format (Figure \ref{fig:prompt}); (2) force the models to generate only the options (e.g. from A to E). 

\subsection{Tuning strategy}
Instruction tuning is needed in the CD evaluation, where instruction templates are required. For instruction-tuned LLMs, we use their own instruction templates; and for foundation models, we use the Alpaca template. Also, we use LoRA \cite{hu2021lora} on the attention blocks to lower the training cost.\footnote{We use the recommended setting in the LLaMA-Factory repository (see Appendix \ref{exp-details}).}
The hyper-parameters and computational resource used in the experiments are listed in Appendix \ref{exp-details}.

\begin{table*}[ht]
    \centering
    \footnotesize
    \begin{tabular}{cccccc}
\hline
Model    & mixed      & cont'      & en                           & zh (/en)                            & others (/en)                        \\ \hline
LLaMA    & \textbf{N} & \textbf{N} & 50.36                        & 15.98 (0.32)                        & 17.95 (0.36)                        \\
Chinese-LLaMA  & N          & Y          & {\color[HTML]{F54A45} 29.38} & {\color[HTML]{34C724} 21.12 (0.72)} & {\color[HTML]{F54A45} 5.88 (0.20)}  \\ \hline
LLaMA2   & \textbf{Y} & \textbf{N} & {\color[HTML]{1F2329} 73.29} & {\color[HTML]{1F2329} 43.57 (0.59)} & 35.78 (0.49)                        \\
Chinese-LLaMA2 & Y          & Y          & {\color[HTML]{F54A45} 60.93} & {\color[HTML]{F54A45} 32.32 (0.53)} & {\color[HTML]{F54A45} 22.59 (0.37)} \\ \hline
LLaMA       & \textbf{N} & \textbf{N} & {\color[HTML]{1F2329} 50.36} & {\color[HTML]{1F2329} 15.98 (0.32)} & {\color[HTML]{1F2329} 17.95 (0.36)} \\
LLaMA2   & Y          & N          & {\color[HTML]{34C724} 73.29} & {\color[HTML]{34C724} 43.57 (0.59)} & {\color[HTML]{34C724} 35.78 (0.49)} \\
Baichuan2-base & Y          & N          & {\color[HTML]{34C724} 87.58} & {\color[HTML]{34C724} 59.30 (0.68)} & {\color[HTML]{34C724} 38.61 (0.44)} \\ \hline
Chinese-LLaMA     & \textbf{N} & \textbf{Y} & {\color[HTML]{1F2329} 29.38} & {\color[HTML]{1F2329} 21.12 (0.72)} & {\color[HTML]{1F2329} 5.88 (0.20)}  \\
Chinese-LLaMA2 & Y          & Y          & {\color[HTML]{34C724} 60.93} & {\color[HTML]{34C724} 32.32 (0.53)} & {\color[HTML]{34C724} 22.59 (0.37)} \\ \hline
\end{tabular}
\caption{The comparison of the selected models' RA scores on the \textit{Basic} knowledge (the mean of xCSQA and xCOPA scores), where "mixed" and "cont'" means having Chinese mixed or continued pretraining, "/en" means the ratio to the English scores, and "others" refers to the mean scores in the other 8 languages. The first lines in each division (LLaMA, LLaMA2, LLaMA and Chinese-LLaMA) are the baseline values (in black). The green values are higher than baseline, and the red ones are lower than baseline. (Same notations in below.)}
\label{tab:pretrain-basic-ra}
\end{table*}

\begin{table*}[ht]
    \centering
    \footnotesize
    \begin{tabular}{cccccc}
\hline
Model    & mixed      & cont'      & en                           & zh (/en)                            & others (/en)                        \\ \hline
LLaMA    & \textbf{N} & \textbf{N} & {\color[HTML]{1F2329} 79.28} & {\color[HTML]{1F2329} 7.49 (0.09)}  & 42.19 (0.53)                        \\
Chinese-LLaMA  & N          & Y          & {\color[HTML]{F54A45} 44.23} & {\color[HTML]{34C724} 13.01 (0.29)} & {\color[HTML]{F54A45} 15.44 (0.35)} \\ \hline
LLaMA2   & \textbf{Y} & \textbf{N} & {\color[HTML]{1F2329} 91.58} & {\color[HTML]{1F2329} 40.39 (0.44)} & 56.00 (0.61)                        \\
Chinese-LLaMA2 & Y          & Y          & {\color[HTML]{F54A45} 79.84} & {\color[HTML]{34C724} 45.92 (0.58)} & {\color[HTML]{F54A45} 43.70 (0.55)} \\ \hline
LLaMA       & \textbf{N} & \textbf{N} & {\color[HTML]{1F2329} 79.28} & {\color[HTML]{1F2329} 7.49 (0.09)}  & {\color[HTML]{1F2329} 42.19 (0.53)} \\
LLaMA2   & Y          & N          & {\color[HTML]{34C724} 91.58} & {\color[HTML]{34C724} 40.39 (0.44)} & {\color[HTML]{34C724} 56.00 (0.61)} \\
Baichuan2-base & Y          & N          & {\color[HTML]{34C724} 86.34} & {\color[HTML]{34C724} 65.81 (0.76)} & {\color[HTML]{34C724} 50.51 (0.59)} \\ \hline
Chinese-LLaMA     & \textbf{N} & \textbf{Y} & {\color[HTML]{1F2329} 44.23} & {\color[HTML]{1F2329} 13.01 (0.29)} & {\color[HTML]{1F2329} 15.44 (0.35)} \\
Chinese-LLaMA2 & Y          & Y          & {\color[HTML]{34C724} 79.84} & {\color[HTML]{34C724} 45.92 (0.58)} & {\color[HTML]{34C724} 43.70 (0.55)} \\ \hline
\end{tabular}
\caption{The comparison of the selected models' RA scores on the \textit{Factual} knowledge (the mean of xGeo and xPeo scores).}
\label{tab:pretrain-realistic-ra}
\end{table*}

\begin{table}[ht]
    \centering
    \footnotesize
    \begin{tabular}{ccccc}
\hline
Model       & mixed      & cont'      & zh                           & others                       \\ \hline
LLaMA       & \textbf{N} & \textbf{N} & {\color[HTML]{1F2329} .8327} & {\color[HTML]{1F2329} .8975} \\
Chinese-LLaMA     & N          & Y          & {\color[HTML]{34C724} .8597} & {\color[HTML]{F54A45} .7764} \\ \hline
LLaMA2  & \textbf{Y} & \textbf{N} & {\color[HTML]{1F2329} .9648} & {\color[HTML]{1F2329} .9498} \\
Chinese-LLaMA2    & Y          & Y          & {\color[HTML]{F54A45} .9536} & {\color[HTML]{F54A45} .9276} \\ \hline
LLaMA       & \textbf{N} & \textbf{N} & {\color[HTML]{1F2329} .8327} & {\color[HTML]{1F2329} .8975} \\
LLaMA2      & Y          & N          & {\color[HTML]{34C724} .9648} & {\color[HTML]{34C724} .9498} \\
Baichuan2-base & Y          & N          & {\color[HTML]{34C724} .9410} & {\color[HTML]{34C724} .9453} \\ \hline
Chinese-LLaMA & \textbf{N} & \textbf{Y} & {\color[HTML]{1F2329} .8597} & {\color[HTML]{1F2329} .7764} \\
Chinese-LLaMA2    & Y          & Y          & {\color[HTML]{34C724} .9536} & {\color[HTML]{34C724} .9276} \\ \hline
\end{tabular}
\caption{The comparison of the selected models' en-CO scores on the \textit{Factual} knowledge (the mean of xGeo and xPeo scores).}
\label{tab:pretrain-realistic-enco}
\end{table}

\begin{table}[ht]
    \centering
    \footnotesize
    \begin{tabular}{ccccc}
\hline
Model          & mixed      & cont'      & zh                           & others                       \\ \hline
LLaMA          & \textbf{N} & \textbf{N} & .0000                        & .0443                        \\
Chinese-LLaMA  & N          & Y          & {\color[HTML]{34C724} .0204} & {\color[HTML]{F54A45} .0436} \\ \hline
LLaMA2         & \textbf{Y} & \textbf{N} & .0153                        & .0570                        \\
Chinese-LLaMA2 & Y          & Y          & {\color[HTML]{F54A45} .0050} & {\color[HTML]{F54A45} .0337} \\ \hline
LLaMA          & \textbf{N} & \textbf{N} & .0000                        & .0443                        \\
LLaMA2         & Y          & N          & {\color[HTML]{34C724} .0153} & {\color[HTML]{34C724} .0570} \\
Baichuan2-base & Y          & N          & {\color[HTML]{1F2329} .0000} & {\color[HTML]{F54A45} .0421} \\ \hline
Chinese-LLaMA & \textbf{N} & \textbf{Y} & {\color[HTML]{1F2329} .0204} & {\color[HTML]{1F2329} .0436} \\
Chinese-LLaMA2 & Y          & Y          & {\color[HTML]{F54A45} .0050} & {\color[HTML]{F54A45} .0337} \\ \hline
\end{tabular}
\caption{The comparison of the selected models' XRR scores on the \textit{Fictional} knowledge.}
\label{tab:pretrain-xrr}
\end{table}

\begin{table}[ht]
    \centering
    \footnotesize
    \begin{tabular}{cccccc}
\hline
Model       & pre        & tune & en     & zh     & others \\ \hline
Alpaca      & \textbf{N} & \textbf{N}    & +25.47 & +6.70  & +9.19  \\
BayLing     & N          & Y    & +25.24 & +22.37 & +8.01  \\ \hline
LLaMA2 Chat & \textbf{Y} & \textbf{N}    & -4.04  & -18.97 & -10.17 \\
Vicuna v1.5 & Y          & Y    & +12.51 & +8.24  & +10.17 \\ \hline
\end{tabular}
\caption{The change in RA scores on the \textit{Basic} knowledge (the mean of xCSQA and xCOPA scores) after instruction tuning, compared with their foundation models (LLaMA and LLaMA2). "pre" and "tune" means whether the model has Chinese pretraining or instruction tuning, and "others" refers to the mean scores in the other 8 languages. (Same notations in below.)}
\label{tab:tune-basic-ra}
\end{table}

\begin{table}[ht]
    \centering
    \footnotesize
    \begin{tabular}{cccccc}
\hline
Model       & pre        & tune       & en     & zh     & others \\ \hline
Alpaca      & \textbf{N} & \textbf{N} & -2.00  & -2.98  & +3.13  \\
BayLing     & N          & Y          & -13.67 & +2.07  & -9.70  \\ \hline
LLaMA2-Chat & \textbf{Y} & \textbf{N} & -10.90 & -15.13 & -6.63  \\
Vicuna v1.5 & Y          & Y          & -7.94  & -11.11 & -2.30  \\ \hline
\end{tabular}
\caption{The change in RA scores on the \textit{Factual} knowledge (the mean of xGeo and xPeo scores) after instruction tuning.}
\label{tab:tune-realistic-ra}
\end{table}

\begin{table}[ht]
    \centering
    \footnotesize
    \begin{tabular}{ccccc}
\hline
Model       & pre        & tune       & zh     & others \\ \hline
Alpaca      & \textbf{N} & \textbf{N} & +.0800 & +.0338 \\
BayLing     & N          & Y          & +.0456 & -.0376 \\ \hline
LLaMA2-Chat & \textbf{Y} & \textbf{N} & -.0203 & +.0081 \\
Vicuna v1.5 & Y          & Y          & +.0116 & +.0054 \\ \hline
\end{tabular}
\caption{The change in en-CO scores on the \textit{Factual} knowledge (the mean of xGeo and xPeo scores) after instruction tuning.}
\label{tab:tune-realistic-enco}
\end{table}

\begin{table}[ht]
    \centering
    \footnotesize
    \begin{tabular}{ccccc}
\hline
Model       & pre        & tune       & zh     & others \\ \hline
Alpaca  & \textbf{N} & \textbf{N} & +.0000 & -.0034 \\
BayLing & N          & Y          & +.0003 & -.0078 \\ \hline
LLaMA2-Chat & \textbf{Y} & \textbf{N} & -.0153 & -.0070 \\
Vicuna v1.5 & Y          & Y          & -.0153 & -.0076 \\ \hline
\end{tabular}
\caption{The change in XRR scores on the \textit{Fictional} knowledge after instruction tuning.}
\label{tab:tune-xrr}
\end{table}

\section{Main Results}
\subsection{General \kalign of multilingual LLMs}
In this part, we assess the basic ability and the \kalign of representative multilingual LLMs among the 10 tested langauges. The findings are:

\paragraph{Basic abilities: imbalanced.} Figure \ref{fig:basic-ra} shows the models' RA scores on the \textit{Basic} knowledge, which reflects the imbalance of basic abilities across different languages, which could be affected by language similarity and resources. For instance, \texttt{en}, \texttt{de}, \texttt{fr}, \texttt{it}, \texttt{pl} and \texttt{ru} belong to the Indo-European family, and they also witness better cross-lingual knowledge alignment with English; On the other hand, \texttt{ar} , \texttt{he}, \texttt{ja} and \texttt{zh} belong to other families, and are non-Latin. Among them, \texttt{ar} and \texttt{he} are also lower-resourced, so it is not surprising that the models show the worst performance on \texttt{ar} and \texttt{he}. Compared with ChatGPT, the LLaMA models and BLOOMZ show larger imbalance.

\paragraph{Factual knowledge alignment: imbalanced PF, but high CT.} Figure \ref{fig:realistic-ra} and \ref{fig:realistic-enco} show the RA and en-CO on the \textit{Factual} knowledge, corresponding to the PF and CT levels of \kalign. One can see the RA scores are also imbalanced, especially in \texttt{zh}, where the factual knowledge performance is too low to match the basic ability. However, the en-CO scores are quite high in all languages, suggesting that the right answers given in non-English languages are very likely the same as English answers.

\paragraph{Factual knowledge alignment: low CD.} There are two possible causes of the high CT: A. The knowledge is conducted from English to non-English; B. The non-English training data is a translated subset of the English training data. Here, our XRR results (Table \ref{tab:general-xrr}) supports the latter. It shows the XRR scores are low across all non-English languages, especially in non-Latin languages. This result suggests that, the high English-CT revealed by the models are more likely an outcome of overlapping training data, instead of knowledge conductivity.

\subsection{Chinese case study on the effect of multilingual pretraining and finetuning}
In this part, we show the effect of multilingual pretraining and instruction tuning by comparing the basic ability and the \kalign of several selected models in Chinese.

\subsubsection{Effect of multilingual pretraining}
\paragraph{Mixed pretraining improves basic abilities, while continued pretraining does not.} Table \ref{tab:pretrain-basic-ra} shows the RA scores on the \textit{Basic} knowledge. Comparing models with and without continued and mixed Chinese pretraining, one can see that mixed pretraining improves the models' basic language abilities in all languages, while continued pretraining has negative effect on them (even in Chinese). This suggest that continued pretraining in a certain language may not be as useful as adding the language in the mixed pretraining process, in order to enhance the model's overall basic language abilities. 

\paragraph{Mixed pretraining improves PF and CT alignment.} Table \ref{tab:pretrain-realistic-ra} and \ref{tab:pretrain-realistic-enco} show the RA and en-CO scores on the \textit{Factual} knowledge. Similar to the \textit{Basic} results, mixed pretraining improves the performance in all languages, as well as enhancing the English consistency of non-English languages. However, continued pretraining in Chinese only improves the Chinese performance, at the cost of lowering performance in other languages. Also, continued pretraining contributes little to the English consistency of non-English languages, including Chinese. This result suggests that, for mixed pretraining, the performance gain is spread in all languages, and the improvement of \kalign is down to the consistency level; However, for continued pretraining, despite the surfacial performance gain in the trained language, it is risky of harming performance in other languages, and does not improve the \kalign in deeper levels.

\paragraph{Multilingual pretraining hardly improves CD alignment.}
Table \ref{tab:pretrain-xrr} shows the XRR scores on the \textit{Fictional} knowledge. One can see that neither continued nor mixed pretraining can bring stable and significant increase to the XRR scores, meaning the knowledge conductivity from English to Chinese is still near zero after Chinese pretraining. This result suggest that current multilingual pretraining methods cannot improve the \kalign in the CD level, which again supported the hypothesis that the high consistency between non-English and English found in current LLMs is an outcome of overlapping training data, not knowldege transfer from English.

\subsubsection{Effect of multilingual finetuning}
\paragraph{Multilingual finetuning improves basic abilities.} Table \ref{tab:tune-basic-ra} shows the RA scores of instruction-tuned LLMs on the \textit{Basic} knowledge. Compared with English-only instruction tuning, adding Chinese data in the tuning process significantly improves the RA scores in Chinese, while not hurting the English performance. This results suggests that multilingual instruction tuning is suitable for fostering basic language abilities in non-English languages.

\paragraph{Multilingual finetuning lowers performance drop in factual knowledge.} Table \ref{tab:tune-realistic-ra} shows the RA scores of instruction-tuned LLMs on the \textit{Factual} knowledge. Surprisingly, both English-only and multilingual instruction tuning causes drop in the RA scores, which indicates a performance drop in factual knowledge after the tuning. Since this phenomenon is not observed on the \textit{Basic} knowledge, this cannot be explained by the "chat bot" preference, but may suggest a shared disadvantage of current instruction tuning strategies. However, compared with English-only tuning, multilingual tuning causes less damage to the factual knowledge performance, contributing to the PF level of \kalign.

\paragraph{Multilingual finetuning can hardly improve CT or CD alignment.} Table \ref{tab:tune-realistic-enco} and \ref{tab:tune-xrr} shows the en-CO scores on the \textit{Factual} knowledge and the XRR scores on the \textit{Fictional} knowledge. One can see that the changes in the two scores brought by English-only and multilingual instructiong tuning are both minor, and multilingual instruction tuning shows no significant advantage over English-only tuning. This result suggests that instruction tuning cannot improve \kalign deeper than the PF level.

\section{Supplement Experiments}
The main results of this paper has shown that the PF and CD levels of current open source LLMs are unsatisfactory, and the CT and CD of them cannot be substantially enhanced by multilingual pretraining or finetuning. However, the result of low CD from English to Chinese can also be due to the linguistic (e.g. lexical, morphological) difference between Chinese and English, or the deficiency of our LoRA finetuning. Thus, adding CD experiments on another Indo-European language and using other finetuning strategies will make our findings more grounded.

\subsection{German case study}
To avoid the effect of low resource on CD, we choose German as a high-resource, Indo-European (Germanic) language to test the models' conductivity to it from English. We adopt a trending German LLM, LeoLM\footnote{\url{https://huggingface.co/LeoLM/leo-hessianai-13b}} (13B, base version), which is based on LLaMA2 and has gone through continued pretraining in German. Also, we compare LLaMA2 and Vicuna v1.5 for having or not having German instruction tuning. (See Appendix \ref{sec:german-case}.)

\paragraph{German continued pretraining harms basic ability and knowledge alignment.}
Table \ref{tab:german-basic-ra} shows the result of LLaMA2 and LeoLM-base on the \textit{Basic} knowledge. Similar to the result of Chinese-LLaMA2, the German continued pretraining of LeoLM leads to the overall decline of basic ability in English, German and other languages. Also, from Table \ref{tab:german-factual-ra}, \ref{tab:german-factual-enco} and \ref{tab:german-fictional-xrr}, we can see the PF, CT and CD levels of \kalign drop for all the tested languages after the continued pretraining. This is consistent with our findings in the Chinese case study.

\paragraph{German finetuning can improve basic ability and knowledge alignment.}
Table \ref{tab:german-tune-basic-ra} shows the result of LLaMA2-chat and Vicuna v1.5 on the \textit{Basic} knowledge, where we can see the multilingual (including German) instruction tuning of Vicuna v1.5 improves the basic ability in German and other languages. Then, from Table \ref{tab:german-tune-factual-ra}, \ref{tab:german-tune-factual-enco} and \ref{tab:german-tune-fictional-xrr}, one can see that it lowers the performance drop in factual knowledge, slightly improves CT, and raises CD. Interestingly, the increase in CD is above chance level (+0.0709 XRR), which is not observed in the Chinese case. This suggest that improving the knowledge conductivity from English to a similar language may be easier than to a less similar one. However, the increased XRR score is only 0.13, which is still not satisfactory for a language like German.

\subsection{Alternative finetuning strategies}
Apart from LoRA tuning on attention blocks only, we add two other experiments using LoRA on all blocks on LLaMA-13B, and fully finetuning on LLaMA 7B. Besides, another experiment that adds extra translation data of the entities from non-English to English is performed to see whether the "reversal curse" \cite{berglund2023reversal} of the unidirectional translation data causes the low XRR results. (See Appendix \ref{sec:alternative-tune}.)

\paragraph{LoRA-all and fully finetuning cannot improve overall CD.} Table \ref{tab:strategy-lora} shows the comparison of LoRA-attention, LoRA-all and fully finetuning on the LLaMA 13B and 7B models. Although the XRR scores are improved in certain languages such as French, Italian and Polish, the overall improvement is minor, and the scores even drops in some other languages. This result shows the low CD results still hold with larger scale finetuning.

\paragraph{Adding reversed translation data cannot improve overall CD.} Table \ref{tab:strategy-reversed} shows the comparison of XRR scores between using unidirectional or bidirectional translation data in the CD experiment on LLaMA2-13B. The results show that adding translation data from non-English to English does not significantly improve the XRR scores, thus the low CD results still hold.

\section{Conclusion}
In this paper, we evaluated the \kalign of representative multilingual LLMs, and systematically assessed the effect of multilingual pretraining and instruction tuning on it, using the proposed \CLKA framework. 

The first part of our results shows that the \kalign of current multilingual LLMs is unsatisfactory, and even though they show high cross-lingual consistency, it is more likely to come from overlapping training data, instead of knowledge conduction between languages. 

The second part of our results demonstrates the effect on basic language ability and knowledge alignment of adding multilingualism in pretraining and instruction tuning, which shows that mixed multilingual pretraining and multilingual instruction tuning is beneficial. However, our results also point out that neither of the two techniques can improve the knowledge conductivity of LLMs, meaning the cross-lingual alignment in current models are still shallow and requires novel strategies for improvement.

\section*{Limitations}
One key limitation of this paper is that the evaluation is restricted to several selected models, which may lead to over-simplification of models in a wider range. Also, the assessment of the effect of multilingual pretraining and instruction tuning only takes English and Chinese into account (and German for supplement), which only covers a narrow set of linguistic features \cite{littell-etal-2017-uriel} and cannot represent the whole picture of multilingual research. These two limitations are partly due to our computational resources and the lack of suitable models for comparison. With adequate resources, our \CLKA framework can be applied to more models and languages to further examine our findings.

Another limitation is that the \textit{Fictional} knowledge requires 2-hop inference to conduct (one for translating the city name, the other for translating the continent name), which is consistent with questions in the xGeo dataset, but it may add too much difficulty of CD alignment, leading to under-estimation of the models' knowledge conductivity. To address this issue, we did a small-scale experiment on LLaMA2-Chat using fictional city names and their founding years. The result shows that the XRR of \texttt{de} rises to 0.26, but that of \texttt{zh} is still very low (around 0.01), which suggests that the low conductivity issue is still existing in questions with lower difficulty.

\section*{Ethics Statement}
The authors declare no competing interests. The datasets used in the evaluation come from publicly available sources and do not contain sensitive contents such as personal information. The adaptation and use of data (CSQA, COPA, Wikidata) are under their licenses. The data generated by ChatGPT other models are non-toxic and used for research only, which is consistent with their intended use.

\section*{Acknowledgements}
We would like to thank the anonymous reviewers for their insightful comments. Shujian Huang is the corresponding author. This work is supported by National Science Foundation of China (No. 62376116, 62176120), the Liaoning Provincial Research Foundation for Basic Research (No. 2022-KF-26-02), research project of Nanjing University-China Mobile Joint Institute.


\appendix

\section{Language choice}
\label{sec:appendix-lang}
Table \ref{tab:language_countries} shows the languages tested in \CLKA.
\begin{table}[ht]
    \centering
    \footnotesize
    \begin{tabular}{ccc}
\hline
ISO & Countries & Langauge Family      \\ \hline
en           & US, UK             & \multirow{2}{*}{Germanic}     \\
de           & Germany, Austria   &                               \\ \hline
fr           & France, Canada     & \multirow{2}{*}{Romance}      \\
it           & Italy              &                               \\ \hline
pl           & Poland             & \multirow{2}{*}{Slavic}       \\
ru           & Russia, Belarus    &                               \\ \hline
ar           & Egypt, Algeria     & \multirow{2}{*}{Afro-Asiatic} \\
he           & Israel             &                               \\ \hline
ja           & Japan              & Japonic                       \\ \hline
zh           & China (Mainland)   & Chinese-Tibetan               \\ \hline
\end{tabular}
    \caption{Correspondence between Languages, Countries, and Language Families}
    \label{tab:language_countries}
\end{table}

\begin{table*}[ht]
    \centering
    \footnotesize
\begin{tabular}{ll}
\hline
Type                 & Templates                                                                       \\ \hline
\multirow{10}{*}{Translation} & Could you convert the upcoming English text to \{lang\}? \{ENTITY\}                                \\
                              & I’d appreciate it if you could transform the following English sentence into \{lang\}.  \{ENTITY\} \\
                     & Please change the following English expression into \{lang\}.  \{ENTITY\}      \\
                     & Kindly rewrite the next English phrase in \{lang\}.  \{ENTITY\}                \\
                     & Can you transmute the subsequent English words into \{lang\}?  \{ENTITY\}      \\
                     & I need the ensuing English to be translated into \{lang\}, please.  \{ENTITY\} \\
                     & Would you mind translating the forthcoming English into \{lang\}?  \{ENTITY\}  \\
                     & Can you render the English text that follows into \{lang\}?  \{ENTITY\}        \\
                     & Please transform the subsequent English language into \{lang\}.  \{ENTITY\}    \\
                              & I require a translation of the upcoming English sentence into \{lang\}, please.  \{ENTITY\}        \\ \hline
\multirow{10}{*}{QA} & Which continent is \{PLACE\} located in?                                       \\
                     & What is the continent of \{PLACE\}?                                            \\
                     & Where is \{PLACE\}? Which continent?                                           \\
                     & In which continent can you find \{PLACE\}?                                     \\
                     & Tell me the continent where \{PLACE\} is located.                              \\
                     & What continent does \{PLACE\} belong to?                                       \\
                     & Where can \{PLACE\} be found continent-wise?                                   \\
                     & What's the continental location of \{PLACE\}?                                  \\
                     & Which part of the world is \{PLACE\} in, continent-wise?                       \\
                     & Could you specify the continent for \{PLACE\}?                                 \\ \hline
\end{tabular}
\caption{Translation and QA templates used to construct the \textit{Fictional} knowledge dataset.}
\label{tab:fictional-templates}
\end{table*}

\begin{figure}[ht]
    \centering
    \includegraphics[width=8cm]{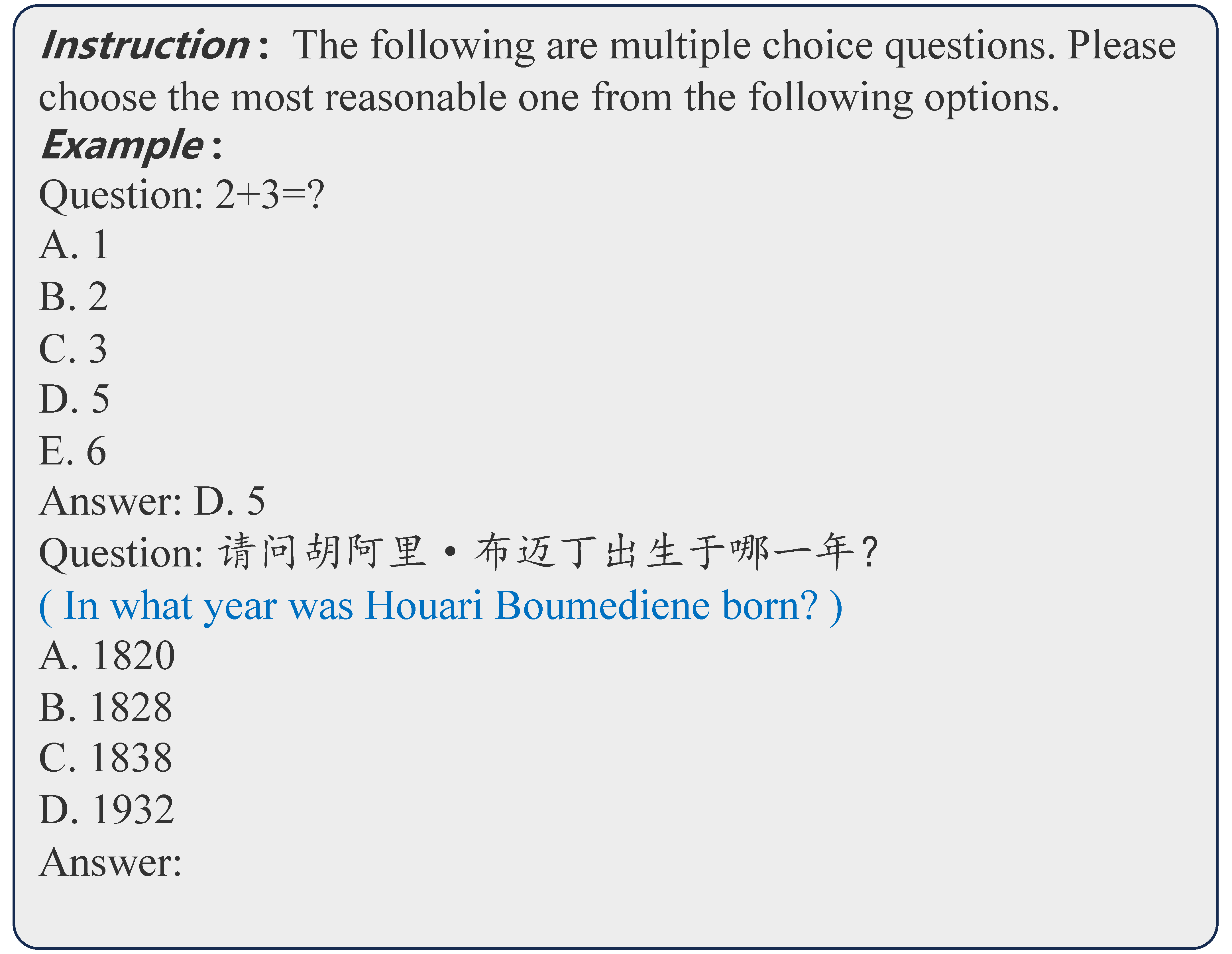}
    \caption{Example prompt for testing models on all our datasets.}
    \label{fig:prompt}
\end{figure}

\begin{figure}[ht]
    \centering
    \includegraphics[width=7.5cm]{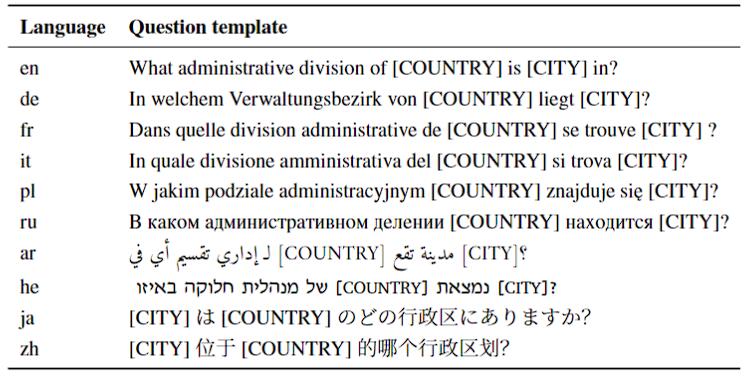}
    \caption{Question templates for the xGeo part of the \textit{Factual} dataset.}
    \label{fig:template-geo}
\end{figure}

\begin{figure*}[ht]
    \centering
    \begin{subfigure}[t]{7.5cm}
        \includegraphics[width=7.5cm]{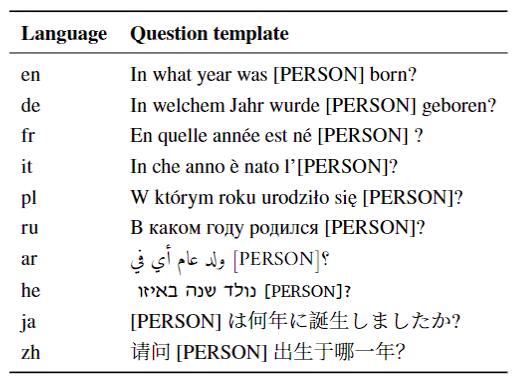}
    \end{subfigure}
    \begin{subfigure}[t]{7.5cm}
        \includegraphics[width=7.5cm]{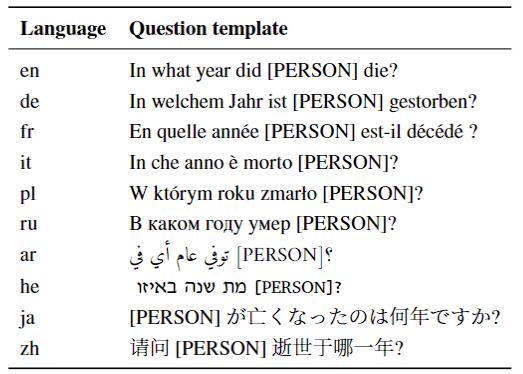}
    \end{subfigure}
    \caption{Question templates for the xPeo part of the \textit{Factual} dataset.}
    \label{fig:template-peo}
\end{figure*}

\section{Example Fictional knowledge}
\label{sec:appendix-example-fictional}
\begin{figure*}
    \centering
    \includegraphics[width=15cm]{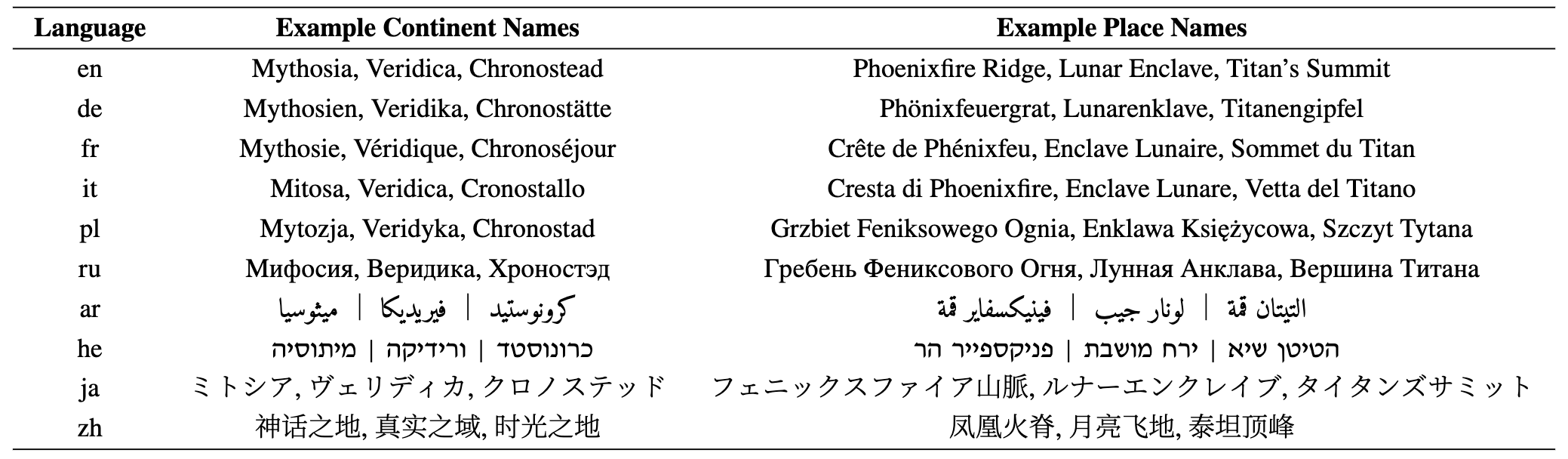}
    \caption{Examples of the continents and places in the \textit{Fictional} data.}
    \label{fig:example-fictional}
\end{figure*}

Figure \ref{fig:example-fictional} shows some example continents and places in the Fictional dataset.

\section{Question templates}
\label{sec:templates}
Figure \ref{fig:prompt} shows the example prompt for testing models in all the datasets. Figure \ref{fig:template-geo} and \ref{fig:template-peo} show the templates we use to construct the \textit{Factual} dataset. Table \ref{tab:fictional-templates} shows the templates we use to construct the \textit{Fictional} dataset.

\section{Experiment details}
\label{exp-details}
This section provides the details in our experiments for replication.

\paragraph{Infrastructure} We use PyTorch and HuggingFace Transformers to load and run the LLMs. For the conductivity experiments, we use the LoRA components in the \href{https://github.com/hiyouga/LLaMA-Factory}{LLaMA-Factory repository} to finetune the models.

\paragraph{Hyper-parameters} For model inference, we set temperature as 0 for ChatGPT and use forced decoding on all other models. For finetuning (in the conductivity experiments), we set training batch size to 32, gradient accumulation steps to 4, training epochs to 3, LoRA rank to 128, LoRA alpha to 16, LoRA dropout to 0.1, and learning rate to 2e-4, which keeps the highest performance of the majority of the models on the \textit{Basic} and \textit{Factual} knowledge after the LoRA finetuning.

\paragraph{LoRA target} The reported CD results in the main body are derived from experiments using LoRA on the attention blocks only. However, we also conducted some experiments using fully finetuning and LoRA on all blocks in the Supplement Experiments.

\paragraph{Computational resources}
All the model inference can be done on 8 Nvidia Tesla v100 32 GB GPUs. Each of the finetuning experiments of 13B models on the \textit{Fictional} knowledge can be done within 2 hours on 4 of those GPUs.

\begin{table}[h]
\centering
\footnotesize
\begin{tabular}{ccccc}
\hline
Model      & mixed      & cont'      & de                           & others                       \\ \hline
LLaMA2     & \textbf{Y} & \textbf{N} & {\color[HTML]{1F2329} .9720} & {\color[HTML]{1F2329} .9498} \\
LeoLM-base & Y          & Y          & {\color[HTML]{F54A45} .9102} & {\color[HTML]{F54A45} .8685} \\ \hline
\end{tabular}
\caption{The en-CO scores of LLaMA2 and LeoLM-base on the \textit{Factual} knowledge (the mean of xGeo and xPeo scores).}
\label{tab:german-factual-enco}
\end{table}

\begin{table}[h]
\centering
\footnotesize
\begin{tabular}{ccccc}
\hline
Model      & mixed      & cont'      & de                           & others                       \\ \hline
LLaMA2     & \textbf{Y} & \textbf{N} & {\color[HTML]{1F2329} .0600} & {\color[HTML]{1F2329} .0515} \\
LeoLM-base & Y          & Y          & {\color[HTML]{F54A45} .0580} & {\color[HTML]{F54A45} .0399} \\ \hline
\end{tabular}
\caption{The XRR scores of LLaMA2 and LeoLM-base on the \textit{Fictional} knowledge.}
\label{tab:german-fictional-xrr}
\end{table}

\begin{table}[h]
\centering
\footnotesize
\begin{tabular}{ccccc}
\hline
Model       & de-tune & en                            & de                            & others                        \\ \hline
LLaMA2-Chat & \textbf{N}       & {\color[HTML]{1F2329} -4.04}  & {\color[HTML]{1F2329} -7.79}  & {\color[HTML]{1F2329} -11.57} \\
Vicuna v1.5 & Y       & {\color[HTML]{373C43} +12.51} & {\color[HTML]{373C43} +20.99} & {\color[HTML]{373C43} +8.57}   \\ \hline
\end{tabular}
\caption{The difference in RA scores after the instruction tuning of LLaMA2-Chat and Vicuna v1.5 on the \textit{Basic} knowledge.}
\label{tab:german-tune-basic-ra}
\end{table}

\begin{table}[h!]
\centering
\footnotesize
\begin{tabular}{ccccc}
\hline
Model       & de-tune & en                            & de                           & others                       \\ \hline
LLaMA2-Chat & \textbf{N}       & {\color[HTML]{1F2329} -10.90} & {\color[HTML]{1F2329} -8.85} & {\color[HTML]{1F2329} -7.42} \\
Vicuna v1.5 & Y       & {\color[HTML]{1F2329} -7.94}  & {\color[HTML]{1F2329} -3.98} & {\color[HTML]{1F2329} +4.23} \\ \hline
\end{tabular}
\caption{The difference in RA scores after the instruction tuning of LLaMA2-Chat and Vicuna v1.5 on the \textit{Factual} knowledge.}
\label{tab:german-tune-factual-ra}
\end{table}

\begin{table*}[h!]
\centering
\footnotesize
\begin{tabular}{ccccccccccc}
\hline
Model     & Strategy & de    & fr    & it    & pl    & ru    & ar    & he    & ja    & zh    \\ \hline
LLaMA-13B & LoRA-Attn     & .0965 & .0950 & .1023 & .0606 & .0000 & .0000 & .0000 & .0000 & .0000 \\
LLaMA-13B &
  LoRA-All &
  {\color[HTML]{F54A45} .0850} &
  {\color[HTML]{34C724} .1850} &
  {\color[HTML]{34C724} .1300} &
  {\color[HTML]{34C724} .0750} &
  {\color[HTML]{34C724} .0050} &
  {\color[HTML]{34C724} .0050} &
  .0000 &
  {\color[HTML]{34C724} .0300} &
  .0000 \\ \hline
LLaMA-7B  & LoRA-Attn     & .1250 & .1450 & .1500 & .1250 & .0150 & .0050 & .0250 & .0000 & .0000 \\
LLaMA-7B &
  Fully &
  {\color[HTML]{F54A45} .0600} &
  {\color[HTML]{34C724} .1950} &
  {\color[HTML]{34C724} .1700} &
  {\color[HTML]{34C724} .1850} &
  {\color[HTML]{F54A45} .0000} &
  {\color[HTML]{F54A45} .0000} &
  {\color[HTML]{F54A45} .0000} &
  .0000 &
  .0000 \\ \hline
\end{tabular}
\caption{The XRR scores measured by different tuning strategies on the \textit{Fictional} knowledge, where "LoRA-Attn" means using LoRA only on the attention blocks, "LoRA-All" means using LoRA on all blocks and "Fully" stands for fully finetuning.}
\label{tab:strategy-lora}
\end{table*}

\begin{table*}[h!]
\centering
\footnotesize
\begin{tabular}{cccccccccc}
\hline
Translation &
  de &
  fr &
  it &
  pl &
  ru &
  ar &
  he &
  ja &
  zh \\ \hline
en-x &
  .0600 &
  .1268 &
  .1100 &
  .1350 &
  .0000 &
  .0000 &
  .0000 &
  .0245 &
  .0153 \\
en-x, x-en &
  {\color[HTML]{34C724} .0806} &
  {\color[HTML]{F54A45} .0800} &
  {\color[HTML]{F54A45} .0950} &
  .1350 &
  .0000 &
  .0000 &
  .0000 &
  {\color[HTML]{F54A45} .0050} &
  {\color[HTML]{F54A45} .0000} \\ \hline
\end{tabular}
\caption{The XRR scores measured with LLaMA2-13B on the \textit{Fictional} knowledge using different translation training, where "en-x" means only translation pairs from English to other languages are provided, and "en-x, x-en" means a equal size of reversed translation data being added using the same templates.}
\label{tab:strategy-reversed}
\end{table*}

\begin{table}[h]
\centering
\footnotesize
\begin{tabular}{cccc}
\hline
Model       & de-tune & de                            & others                        \\ \hline
LLaMA2-Chat & \textbf{N}       & {\color[HTML]{1F2329} -.0075} & {\color[HTML]{1F2329} +.0065} \\
Vicuna v1.5 & Y       & {\color[HTML]{373C43} +.0022} & {\color[HTML]{373C43} +.0066} \\ \hline
\end{tabular}
\caption{The difference in en-CO scores after the instruction tuning of LLaMA2 and Vicuna v1.5 on the \textit{Factual} knowledge.}
\label{tab:german-tune-factual-enco}
\end{table}

\begin{table}[h!]
\centering
\footnotesize
\begin{tabular}{cccc}
\hline
Model       & de-tune & de                            & others                        \\ \hline
LLaMA2-Chat & \textbf{N}       & {\color[HTML]{1F2329} +.0200} & {\color[HTML]{1F2329} -.0114} \\
Vicuna v1.5 & Y       & {\color[HTML]{373C43} +.0709} & {\color[HTML]{373C43} -.0183} \\ \hline
\end{tabular}
\caption{The difference in XRR scores after the instruction tuning of LLaMA2 and Vicuna v1.5 on the \textit{Fictional} knowledge.}
\label{tab:german-tune-fictional-xrr}
\end{table}

\section{Introduction of the tested models}
\label{sec:appendix-intro-models}
This section introduces the models used in this research. The model parameter sizes are all 13B unless specified.
\paragraph{Foundation models.} We use the following foundation models:
\begin{itemize}
    \item LLaMA 1\&2 \cite{touvron2023llama, touvron2023llama2}. They are pretrained on mainly English data and a small portion of non-English data. For LLaMA 1, the multilingual pretraining data is basically Wikipedia ($4.5\%$ of the total 1.4T tokens) in 20 languages, including \texttt{en}, \texttt{fr}, it, \texttt{fr}, po and \texttt{ru}. For LLaMA 2, the multilingual data are extended both in quantity and language coverage (\texttt{zh} and \texttt{ja} are added). We use the 70B and 13B versions of LLaMA 2.
    \item Chinese-LLaMA 1\&2 \cite{cui2023efficient}. They are built on LLaMA 1 and 2 respectively, with vocabularies and tokenizers adapted for Chinese, and continued pretraining on 120GB Chinese data. \footnote{We use the "plus" version of Chinese-LLaMA and the "pro" version of Chinese-Alpaca since they are recommended on the \href{https://github.com/ymcui/Chinese-LLaMA-Alpaca}{GitHub page}.}
    \item Baichuan 2-Base \cite{yang2023baichuan}. It is pretrained on 2.6T tokens of multilingual, especially English-Chinese bilingual data.
    \item BLOOM \cite{workshop2023bloom}. It is a foundation model pretrained on 46 languages and 13 programming languages, including \texttt{en}, \texttt{fr}, \texttt{zh} and \texttt{ar}. We use the 7.1B version of it.
\end{itemize}

\begin{table*}[!h]
\centering
\footnotesize
\begin{tabular}{cccccc}
\hline
Model      & mixed & cont' & en                           & de (/en)                            & others (/en)                        \\ \hline
LLaMA2     & Y     & N     & {\color[HTML]{1F2329} 73.29} & {\color[HTML]{1F2329} 46.19 (0.63)} & {\color[HTML]{1F2329} 35.45 (0.48)} \\
LeoLM-base & Y     & Y     & {\color[HTML]{F54A45} 61.10} & {\color[HTML]{F54A45} 41.44 (0.68)} & {\color[HTML]{F54A45} 22.13 (0.36)} \\ \hline
\end{tabular}
\caption{The RA scores of LLaMA2 and LeoLM-base on the \textit{Basic} knowledge (the mean of xCSQA and xCOPA scores), where "mixed" and "cont'" means having German mixed or continued pretraining, "/en" means the ratio to the English scores, and "others" refers to the mean scores in the other 8 languages.}
\label{tab:german-basic-ra}
\end{table*}

\begin{table*}[!h]
\centering
\footnotesize
\begin{tabular}{cccccc}
\hline
Model      & mixed & cont' & en                           & de (/en)                            & others (/en)                        \\ \hline
LLaMA2     & Y     & N     & {\color[HTML]{1F2329} 91.58} & {\color[HTML]{1F2329} 82.52 (0.90)} & {\color[HTML]{1F2329} 50.73 (0.55)} \\
LeoLM-base & Y     & Y     & {\color[HTML]{F54A45} 48.56} & {\color[HTML]{F54A45} 41.91 (0.86)} & {\color[HTML]{F54A45} 18.68 (0.38)} \\ \hline
\end{tabular}
\caption{The RA scores of LLaMA2 and LeoLM-base on the \textit{Factual} knowledge (the mean of xGeo and xPeo scores).}
\label{tab:german-factual-ra}
\end{table*}

\paragraph{Instruction-tuned LLMs.} We use the following instruction tuned models:
\begin{itemize}
    \item Stanford Alpaca \cite{alpaca}. It is LLaMA tuned with 52K English instruction data, and shows improved performance on several LLM benchmarks such as MMLU.
    \item Vicuna \cite{vicuna2023}. We use the v1.5 version of it, which is LLaMA 2 tuned with 70K user conversations with ChatGPT, collected from the ShareGPT website. Because the data is shared by users worldwide, it contains multilingual instructions in various languages.
    \item BayLing \cite{zhang_bayling_2023}. It is LLaMA tuned on interactive translation data and instruction data in \texttt{en}, \texttt{de} and \texttt{zh}. It is reported to show high translation and instruction-following performance on these languages.
    \item Chinese-Alpaca 1\&2 \cite{cui2023efficient}. They are the Chinese-LLaMAs added Alpaca-style \cite{alpaca} instruction tuning. The tuning data is bilingual in English and Chinese. They show much improved performance in Chinese compared with the LLaMA models.
    \item LLaMA 2-Chat \cite{touvron2023llama2}. It is LLaMA 2 with instruction tuning and RLHF. Although not clearly stated, the data used in the finetuning process is inferred to be mainly in English. It shows chatting ability in multiple languages.
    \item Baichuan 2-Chat \cite{yang2023baichuan}. It is Baichuan 2-Base undergone instruction tuning and reinforcement learning. The training is also multilingual, especially bilingual in English and Chinese.
    
    
    \item BLOOMZ-MT \cite{workshop2023bloom}. BLOOMZ-MT is BLOOM tuned on the xP3 instruction dataset \cite{muennighoff-etal-2023-crosslingual} in 46 languages and translation data in 9 languages. The two datasets covers \texttt{en}, \texttt{fr}, \texttt{zh}, \texttt{ar}, \texttt{de} and \texttt{ru}.
\end{itemize}

\section{Results of the German case study}
\label{sec:german-case}
For multilingual pretraining, Table \ref{tab:german-basic-ra} shows the basic ability of LLaMA2 and LeoLM-base; Table \ref{tab:german-factual-ra} and \ref{tab:german-factual-enco} show their PF and CT alignment on the \textit{Factual} knowledge; and Table \ref{tab:german-fictional-xrr} shows their CD alignment on the \textit{Fictional} knowledge.

For multilingual finetuning, Tabel \ref{tab:german-tune-basic-ra} shows the basic ability of LLaMA2-Chat and Vicuna v1.5; Table \ref{tab:german-tune-factual-ra} and \ref{tab:german-tune-factual-enco} show their PF and CT alignment on the \textit{Factual} knowledge; and Table \ref{tab:german-tune-fictional-xrr} shows their CD alignment on the \textit{Fictional} knowledge.

\section{Results of the alternative finetuning strategies}
\label{sec:alternative-tune}
Table \ref{tab:strategy-lora} shows the CD results of the LLaMA models using different tuning techniques; and Table \ref{tab:strategy-reversed} shows the comparison of adding or not adding reversed translation data in the tuning process on the LLaMA2 model.

\end{document}